\pdfoutput=1

\documentclass[11pt]{article}

\usepackage[final]{acl}

\usepackage{times}
\usepackage{latexsym}

\usepackage[T1]{fontenc}

\usepackage[utf8]{inputenc}

\usepackage{microtype}

\usepackage{inconsolata}

\usepackage{graphicx}

\usepackage{multirow}
\usepackage{makecell} 
\usepackage{tabularx}
\usepackage{booktabs}
\usepackage{graphicx}     
\usepackage{rotating} 
\usepackage{array}
\usepackage{url}
\usepackage{placeins} 
\usepackage{longtable}
\usepackage{hyperref}   

\title{GASE: Generatively Augmented Sentence Encoding}

\author{
    Manuel Frank \\
    Department of Computer Science \\
    Munster Technological University \\
    \texttt{Manuel.Frank@zohomail.eu} \\\And
    Haithem Afli \\
    Department of Computer Science \\
    Munster Technological University \\
    \texttt{Haithem.Afli@mtu.ie} \\
    }

\begin{document}
\maketitle
\begin{abstract}
We propose a training-free approach to improve sentence embeddings leveraging test-time compute by applying generative text models for data augmentation at inference time. Unlike conventional data augmentation that utilises synthetic training data, our approach does not require access to model parameters or the computational resources typically required for fine-tuning state-of-the-art models. Generatively Augmented Sentence Encoding variates the input text by paraphrasing, summarising, or extracting keywords, followed by pooling the original and synthetic embeddings. Experimental results on the Massive Text Embedding Benchmark for Semantic Textual Similarity (STS) demonstrate performance improvements across a range of embedding models using different generative models for augmentation. We find that generative augmentation leads to larger performance improvements for embedding models with lower baseline performance. These findings suggest that integrating generative augmentation at inference time adds semantic diversity and can enhance the robustness and generalisability of sentence embeddings for embedding models. Our results show that performance gains depend on the embedding model and the dataset.
\end{abstract}
\begin{figure}[t]
  \centering
  \includegraphics[width=\linewidth]{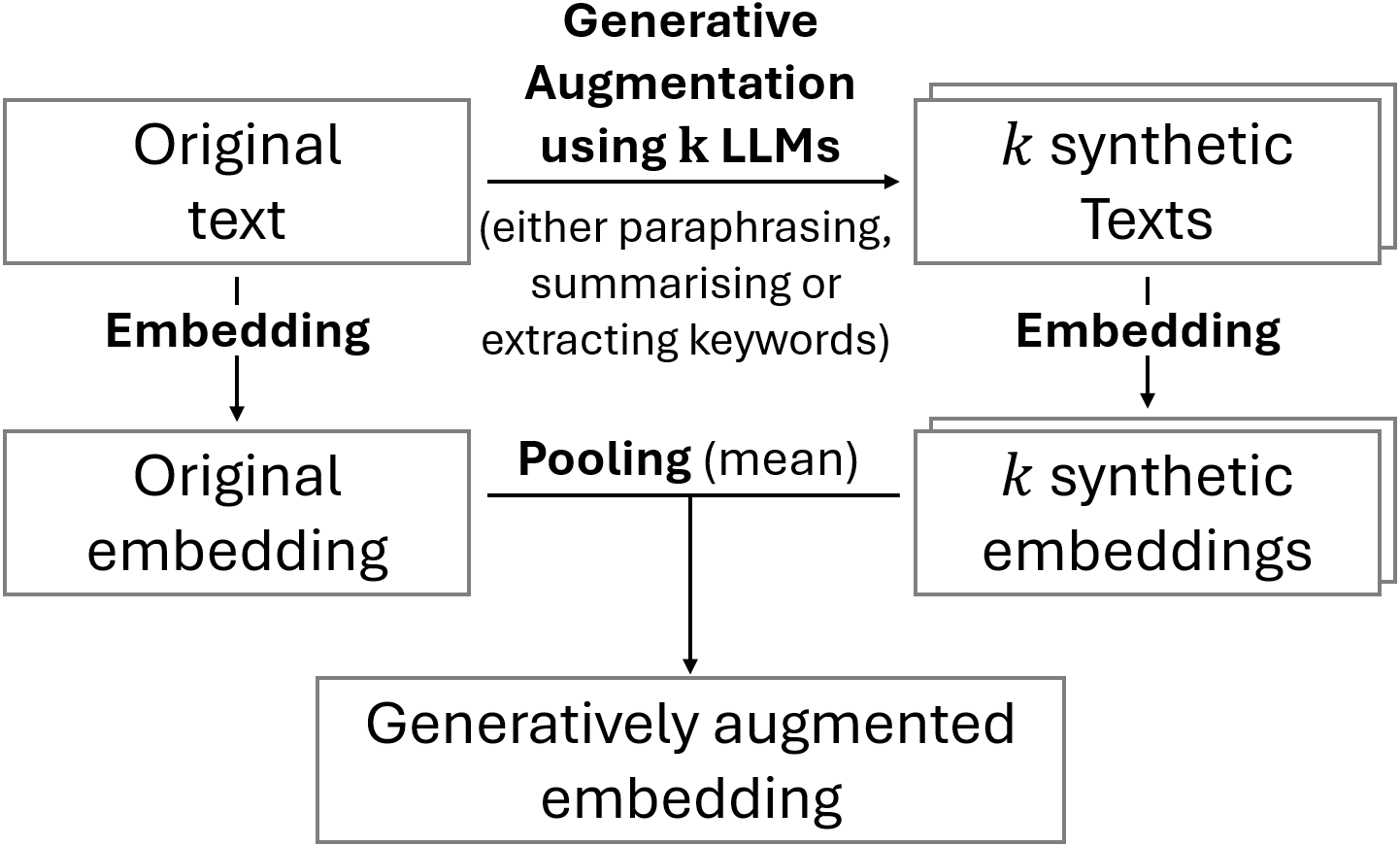}
  \caption{Approach for Generatively Augmented Sentence Encoding.}
  \label{fig:approach}
\end{figure}
\section{Introduction}
\begin{figure}[t]
  \centering
  \includegraphics[width=\linewidth]{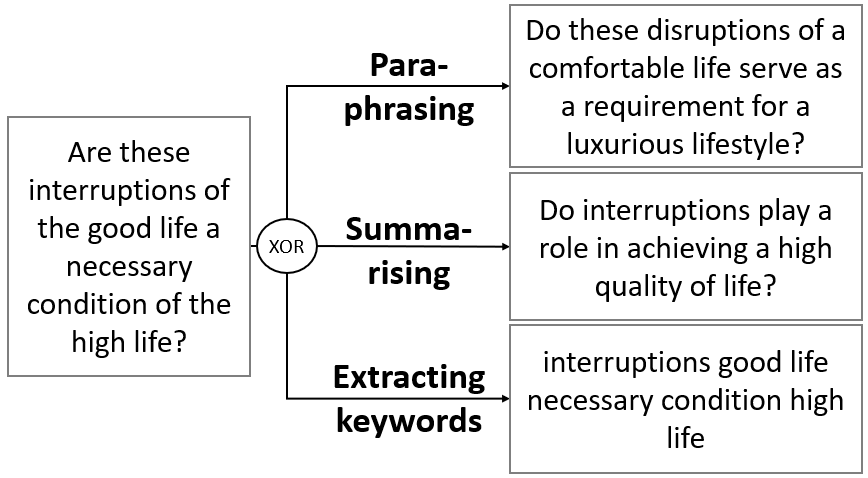}
  \caption{Augmentation examples for paraphrasing, summarising, and extracting keywords.}
  \label{fig:example}
\end{figure}
%
\begin{table*}
\small
\renewcommand{\arraystretch}{1.2} 
\centering
\begin{tabular}{l cc c cc c cc c cc}\hline
\textbf{Embedding model} & \multicolumn{2}{c}{\textbf{No augmentation}} & & \multicolumn{2}{c}{\textbf{Paraphrasing}} & & \multicolumn{2}{c}{\textbf{Summarising}} & & \multicolumn{2}{c}{\textbf{\begin{tabular}[c]{@{}c@{}}Extracting keywords\end{tabular}}} \\\cline{2-3}\cline{5-6}\cline{8-9}\cline{11-12}
& \textbf{Avg.} & \textbf{STS22} & & \textbf{Avg.} & \textbf{STS22} & & \textbf{Avg.} & \textbf{STS22} & & \textbf{Avg.} & \textbf{STS22} \\\hline
glove.840B.300d         & 57.86 & 54.08 & & 59.98 & 57.78 & & 60.64 & \textbf{61.21} & & \textbf{60.82} & 56.94 \\
bert-large-cased        & 60.78 & 55.48 & & 66.29 & 58.57 & & \textbf{66.89} & \textbf{63.47} & & 62.62 & 54.64 \\
all-MiniLM-L6-v2        & 78.91 & 67.26 & & \textbf{80.32} & 68.47 & & 79.90 & \textbf{69.27} & & 77.96 & 64.57 \\
all-mpnet-base-v2       & 80.28 & 68.00 & & 81.39 & 69.01 & & \textbf{81.48} & \textbf{70.12} & & 79.95 & 66.19 \\
embed-english-light-v3.0& 78.75 & 67.88 & & \textbf{80.30} & 68.42 & & 80.21 & \textbf{68.97} & & 78.53 & 67.27 \\
embed-english-v3.0      & 81.25 & 68.22 & & \textbf{82.10} & 68.63 & & 82.08 & \textbf{70.29} & & 81.26 & 66.91 \\
voyage-2                & 82.50 & 65.26 & & \textbf{83.38} & 66.80 & & 82.74 & \textbf{68.50} & & 81.21 & 65.29 \\
voyage-lite-02-instruct & 85.95 & 78.62 & & \textbf{85.99} & 79.08 & & 85.43 & \textbf{79.87} & & 83.91 & 76.60 \\
voyage-large-2          & 83.60 & 63.97 & & \textbf{83.92} & 64.69 & & 83.66 & \textbf{65.90} & & 81.98 & 63.68 \\
voyage-large-2-instruct & 84.61 & 66.59 & & \textbf{84.80} & 67.72 & & 84.54 & \textbf{68.66} & & 83.40 & 65.73 \\
mxbai-embed-large-v1    & 84.63 & 68.73 & & \textbf{85.01} & 70.03 & & 84.45 & \textbf{70.36} & & 83.42 & 67.83 \\
llama-3-8b            & 71.36 & 33.39 & & 75.28 & 50.55 & & \textbf{76.74} & \textbf{59.44} & & 74.95 & 51.79 \\ \hline
\end{tabular}
\caption{Average scores across MTEB STS datasets and STS22 individually for all embedding models with and without generative augmentation using GPT-3.5 Turbo in \%, bold: highest average scores for average and STS22 respectively).}
\label{tab:scores_overview}
\end{table*}
%
\begin{table}
\small
\renewcommand{\arraystretch}{1.2} 
\centering
\begin{tabular}{lccc}\hline
\textbf{Dataset} & 
\makecell{\textbf{No Aug-} \\ \textbf{-mentation}} & 
\makecell{\textbf{Keyword} \\ \textbf{Extraction}} & 
\makecell{\textbf{Random} \\ \textbf{Keyword} \\ \textbf{Extraction}} \\ \hline
STSB              & 50.73           & \textbf{59.26}     & 23.95 \\
STS12             & 57.50           & \textbf{58.09}     & 34.78 \\
STS13             & 70.98           & \textbf{71.12}     & 22.25 \\
STS14             & \textbf{60.69}  & 60.34             & 28.48 \\
STS15             & 70.85           & \textbf{72.44}     & 33.86 \\
STS16             & 63.85           & \textbf{65.31}     & 24.99 \\
STS17             & 62.05           & \textbf{75.92}     & 36.47 \\
STS22             & 54.08           & \textbf{56.94}     & 38.44 \\
SICK-R            & \textbf{55.42}  & 55.34             & 40.04 \\
BIOSSES           & 32.45           & \textbf{33.50}     & 29.02 \\ \hline
\textbf{Average}  & 57.86           & \textbf{60.82}     & 31.23 \\ \hline
\end{tabular}
\caption{STS scores per dataset for GloVe embeddings without generative augmentation vs. generative augmentation using LLM-extracted keywords with GPT-3.5 Turbo or random keyword extraction (Spearman's rank correlation in \%, bold: highest scores).}
\label{tab:scores_rke}
\end{table}
%
%
\begin{table}
\small
\centering
\renewcommand{\arraystretch}{1.1}
\begin{tabular}{@{}lllll@{}} 
\hline
\textbf{Dataset} & \textbf{Original} & \textbf{\begin{tabular}[c]{@{}l@{}}Para-\\ phrases\end{tabular}} & \textbf{\begin{tabular}[c]{@{}l@{}}Summa-\\ ries\end{tabular}} & \textbf{\begin{tabular}[c]{@{}l@{}}Extracted\\ keywords\end{tabular}} \\ \hline
STSB             & 10.1                                                                 & 11.5                                                              & 10.7                                                             & 5.7                                                                    \\
STS12            & 11.1                                                                 & 11.6                                                              & 11.1                                                             & 6.1                                                                    \\
STS13            & 9.0                                                                  & 10.5                                                              & 11.1                                                             & 5.4                                                                    \\
STS14            & 9.3                                                                  & 11.0                                                              & 11.4                                                             & 5.7                                                                    \\
STS15            & 10.6                                                                 & 12.0                                                              & 11.5                                                             & 5.7                                                                    \\
STS16            & 11.6                                                                 & 12.6                                                              & 12.8                                                             & 5.7                                                                    \\
STS17            & 8.7                                                                  & 9.6                                                               & 9.0                                                              & 4.5                                                                    \\
STS22            & 477.2                                                                & 216.5                                                             & 58.4                                                             & 103.8                                                                  \\
SICK-R           & 9.6                                                                  & 10.0                                                              & 9.3                                                              & 4.7                                                                    \\
BIOSSES          & 24.5                                                                 & 25.2                                                              & 19.2                                                             & 13.9                                                                   \\ \hline
\textbf{Average} & 58.2                                                                 & 33.1                                                              & 16.4                                                             & 16.1                                                                   \\
\hline
\end{tabular}
\caption{Average word count using GPT-3.5 Turbo for paraphrasing, summarising and extracting keywords.}
\label{tab:word_count_augmentation}
\end{table}
\begin{table}
\small
\renewcommand{\arraystretch}{1.2}
\centering
\begin{tabular}{lll}\hline
\textbf{Dataset}       & \textbf{Paraphrases} & \textbf{Summaries} \\\hline
STSB           & 0.39                 & 0.52               \\
STS12                  & 0.39                 & 0.49               \\
STS13                  & 0.33                 & 0.41               \\
STS14                  & 0.37                 & 0.47               \\
STS15                  & 0.42                 & 0.53               \\
STS16                  & 0.37                 & 0.40               \\
STS17                  & 0.41                 & 0.59               \\
STS22                  & 0.39                 & 0.21               \\
SICK-R                 & 0.46                 & 0.67               \\
BIOSSES                & 0.46                 & 0.52               \\\hline

\textbf{Average} & 0.40        & 0.48       \\\hline
\end{tabular}
\caption{Average Jaccard Similarity between original texts and paraphrases and summaries generated with GPT-3.5 Turbo respectively.}
\label{tab:jaccard_sim}
\end{table}
\begin{table*}
\small
\renewcommand{\arraystretch}{1.2} 
\centering
\begin{tabular}{lccc|c}\hline
\textbf{Augmentation} & \makecell{\textbf{Original text 1 vs} \\ \textbf{Augmented text 1}} & \makecell{\textbf{Original text 2 vs} \\ \textbf{Augmented text 2}} & \makecell{\textbf{Augmented text 1 vs} \\ \textbf{Augmented text 2}} & \makecell{\textbf{Original text 1 vs} \\ \textbf{Original text 2}} \\\hline
Paraphrasing          & 85.44 & 85.24 & 57.31 & \multirow{3}{*}{\makecell[c]{$\uparrow$\\ 59.81 \\ $\downarrow$}} \\
Summarising           & 85.53 & 85.36 & 57.60 & \\
Keyword Extraction              & 84.27 & 84.20 & 59.81 & \\\hline
\end{tabular}
\caption{Average cosine similarity between original text pairs and augmented texts based on all-mpnet-base-v2 embeddings in \%. (The last column shows cosine similarity between the original, non-augmented text pairs.)}
\label{tab:augmentation_similarity}
\end{table*}
Representation learning has emerged as a fundamental technique in natural language processing (NLP). However, the quality and robustness of the embeddings are highly dependent on the richness and diversity of the training data. Recent advancements of generative Large Language Models (LLMs) led to remarkable capabilities in generating human-like text. LLMs were also used for data augmentation by \citet{Dai2023}, who applied paraphrasing techniques to train a BERT model~\citep{Devlin2019}. \citet{Wahle2022} showed that LLM-generated paraphrases are harder to detect for humans than paraphrases generated with simple techniques like synonym replacement. The integration of generative models with representation learning has been explored by augmenting generative models, e.g., in Retrieval-Augmented Generation~\citep{Lewis2020, Guu2020}. 

In contrast, we introduce an approach to augment embedding models that applies generative models using test-time compute: Generatively Augmented Sentence Encoding (GASE). Instead of generating synthetic training data, GASE creates textual variants of an input text through \textit{paraphrasing}, \textit{summarising}, or \textit{extracting keywords} at inference time. A joint embedding is derived by pooling the embeddings of the original text and the generated transformation(s). The underlying hypothesis of our work is that adding textual diversity using generative models increases the ability of current embedding models to model semantics and hence benefits the performance of downstream STS tasks.

Also aiming to propose a training-free method, \citet{Lei2024} introduced Meta-Task Prompting, which employs use-case-specific prompts to generate multiple embeddings via a generative model, after which the embeddings are pooled. For retrieval systems, \citet{Gao2023} and \citet{Wang2023} described query expansion approaches using generative models that generate synthetic documents as a response to a retrieval request. Extending this work, GASE combines generative and embedding models without any assumptions on the type of models (unlike Meta-Task Prompting), nor is it limited to augmentation of retrieval queries such as \citet{Gao2023} and \citet{Wang2023}. More recently, GenEOL~\citep{Thirukovalluru2024}, a related approach, was introduced (see \autoref{tab:app_gase_geneol} in the Appendix for a comparison with GASE). 
\section{Method}

 GASE performs the following steps (see \autoref{fig:approach}):

\textbf{Generative Augmentation.} We apply $k$ different generative models\footnote{Using a single generative model for $k=2$ did not yield textually diverse and meaningful variations.} to produce $k$ variations of the original input sequence using exactly one of the following transformations: \textit{Paraphrasing}, which provides a semantically equivalent but lexically or syntactically different variation of the input text. \textit{Summarising}, which produces a shorter output text that captures the most important information of the input text. \textit{Extracting Keywords}, which lists the most relevant words from a given text. \autoref{fig:example} provides an example for each. We evaluated $k \in \{0, 1, 2, 3\}$ using GPT-3.5 Turbo~\citep{GPT-3-5-Turbo}, Reka-Flash~\citep{RekaModelOverview}, and GPT-4o mini ~\citep{GPT-4o-mini}\footnote{For model versions and hyperparameters see \autoref{sec:appendix-implementation-details}.}. 

\textbf{2. Sentence Embedding.} Generating embeddings for the original and $k$ synthetic texts with one of $12$ encoders.\footnote{See \mbox{\autoref{sec:appendix-embedding-models}} for a list incl. references.}

\textbf{3. Pooling.} The embeddings of the original text and the $k$ synthetic texts are pooled by computing their arithmetic mean.\footnote{Mean pooling dominates max pooling (see \autoref{tab:pooling_scores_avg}).}

We evaluate our approach on the English subtasks of the MTEB STS task\footnote{See \autoref{sec:appendix-datasets} for details on the datasets.} 
using cosine similarity and Spearman's rank correlation\footnote{See ~\citet{Reimers2016} for why Spearman's rank correlation is preferable over Pearson correlation for STS tasks.}.

\section{Results}
%
%
\begin{table*}
\small
\centering
\renewcommand{\arraystretch}{1.1}
\begin{tabular}{l c c c c c c}
\hline
\textbf{Embedding model} & \textbf{No Aug- } & \textbf{gpt-3.5-turbo} & \textbf{reka-flash} & \textbf{gpt-4o-mini} & \textbf{gpt-3.5-turbo} & \textbf{gpt-3.5-turbo}\\
& \textbf{mentation} & & & & \textbf{+ reka-flash} & \textbf{+ reka-flash}\\
& & & & & &  \textbf{+ gpt-4o-mini}\\\hline
glove.840B.300d & 57.86 & 59.98 & \underline{60.73} & 60.04 & 61.97 & \textbf{62.18} \\
bert-large-cased & 60.78 & \underline{66.29} & 65.76 & 65.01 & 68.01 & \textbf{68.41} \\
all-MiniLM-L6-v2 & 78.91 & \underline{80.32} & 79.74 & 79.81 & 80.64 & \textbf{80.76} \\
all-mpnet-base-v2 & 80.28 & \underline{81.39} & 81.01 & 81.07 & 81.72 & \textbf{81.85} \\
embed-english-light-v3.0 & 78.75 & \underline{80.30} & 79.72 & 79.90 & 80.54 & \textbf{80.66} \\
embed-english-v3.0 & 81.25 & \underline{82.10} & 82.00 & 81.82 & \textbf{82.53} & 82.08 \\
voyage-2 & 82.50 & \underline{83.38} & 83.37 & 83.07 & \textbf{83.80} & 83.72 \\
voyage-lite-02-instruct & 85.95 & \underline{85.99} & 85.92 & 85.97 & \textbf{86.05} & 86.01 \\
voyage-large-2 & 83.60 & 83.92 & \underline{83.98} & 83.95 & 84.12 & \textbf{84.17} \\
voyage-large-2-instruct & 84.61 & \underline{84.80} & 84.68 & 84.78 & 84.86 & \textbf{84.91} \\
mxbai-embed-large-v1 & 84.63 & \underline{85.01} & 84.80 & 84.90 & \textbf{85.17} & 85.16 \\
llama-3-8b & 71.36 & 75.28 & \underline{77.02} & 74.71 & \textbf{77.85} & 77.52 \\ \hline
\end{tabular}
\caption{Average STS scores without and with paraphrase augmentation
using different generative models in \% (bold: highest score per emb. model, underlined: highest score for $k<=1$ per emb. model).}
\label{tab:scores_paraphrase_per_model}
\end{table*}
\autoref{tab:scores_overview} shows the results for different augmentation strategies vs. the respective baseline (see \autoref{tab:app_scores_per_augm_details} in the Appendix for the results per dataset). All embedding models improved their avg. scores through augmentation with the greatest improvements for Llama-3 (+$6.49$pp), BERT (+$6.11$pp), and GloVe (+$2.96$pp). Paraphrase augmentation worked best for most models; however, GloVe improved most with keyword extraction, while BERT, MPNET, and Llama-3 excelled with summarisation.

To validate our approach we compared the augmentation with Keyword Extraction using LLMs to randomly extracting keywords\footnote{See \autoref{sec:appendix-random-keyword-extraction} for implementation details.} which demonstrated that LLMs extract semantically more meaningful keywords (\autoref{tab:scores_rke}).\footnote{Additionally, we compared LLM-based keyword extraction with stopword removal showing that it generates semantically richer extractions (see \mbox{\autoref{tab:app_scores_stopword}} in the Appendix).}

Summarising and extracting keywords significantly reduced the average word count across datasets (see \autoref{tab:word_count_augmentation}). However, GPT-3.5 Turbo could not maintain STS22's text length when paraphrasing and at the same time only effectively summarised STS22. Comparing the Jaccard Similarity between original texts and paraphrases on the one hand and original texts and summaries on the other hand, \autoref{tab:jaccard_sim} shows that paraphrases deviate more from the original texts on all datasets except STS22. At the semantic level (quantified by cosine similarity between embeddings) all augmentation strategies preserve meaning: similarities between the transformed and original texts range from $84.20\%$ to $85.36\%$ (\autoref{tab:augmentation_similarity}).

Evaluating the different generative models for paraphrase augmentation we find that using a single generative model GPT-3.5 Turbo performed best for all embedding models, except GloVe, Voyage-Large-2, and Llama-3 (\autoref{tab:scores_paraphrase_per_model}).\footnote{For results per dataset see \autoref{tab:app_paraphrasing_per_model_details_1} and \autoref{tab:app_paraphrasing_per_model_details_2} in the Appendix).} 

Using weighted averages between paraphrases and originals peaked at 25\%/75\% (excl. BERT) and was worst with paraphrases only (\autoref{fig:weight_analysis}).

Applying more than one generative model for augmentation yielded higher scores than a single model (see \autoref{tab:scores_paraphrase_per_model}) with BERT (+$7.63$pp), Llama-3 (+$6.49$pp) and GloVe (+$4.32$pp) showing the largest gains.
Moreover, the overall performance of embedding models with lower non-augmented performance was associated with larger performance improvements through generative augmentation.
%
\begin{table}
\small
\centering
\renewcommand{\arraystretch}{1.1} 
\begin{tabular}{llcc}
\hline
\textbf{Embedding model} & \textbf{Augment.} & \textbf{STS17} & \textbf{STS22} \\ \hline

\multirow{3}{*}{\makecell[l]{bert-base-multi-\\lingual-cased}}
    & None        & 37.16            & 29.50            \\
    & Paraphrase  & 43.93            & 30.36            \\
    & Summarise   & \textbf{47.13}   & \textbf{42.03}   \\ \hline

\multirow{3}{*}{\makecell[l]{paraphrase-multiling-\\ual-mpnet-base-v2}}
    & None        & 82.51            & 61.56            \\
    & Paraphrase  & \textbf{82.95}   & 62.72            \\
    & Summarise   & 82.88            & \textbf{66.70}   \\ \hline

\multirow{3}{*}{\makecell[l]{multilingual-\\e5-large-instruct}}
    & None        & 83.39            & 69.64            \\
    & Paraphrase  & \textbf{83.95}   & 66.48            \\
    & Summarise   & 83.60            & \textbf{70.26}   \\ \hline

\multirow{3}{*}{\makecell[l]{voyage-multi-\\lingual-2}}
    & None        & 85.14            & 69.88            \\
    & Paraphrase  & \textbf{85.73}   & 70.21            \\
    & Summarise   & 85.41            & \textbf{71.92}   \\ \hline

\end{tabular}
\caption{Average STS scores for multilingual datasets with and without augmentation with GPT-4o mini in \% (bold: highest scores per embedding model and dataset).}
\label{tab:scores_multilingual}
\end{table}

\begin{table}[h]
\centering
\small
\renewcommand{\arraystretch}{1.1}

\begin{tabularx}{\linewidth}{@{}l*{5}{>{\centering\arraybackslash}X}@{}}
\hline
\multicolumn{6}{c}{\textbf{Generative Model Size (Qwen2.5)}} \\ 
\hline
\textbf{None}
& \textbf{0.5b}
& \textbf{1.5b}
& \textbf{3b}
& \textbf{7b}
& \textbf{14b} \\\hline 
80.28
& 79.20
& 79.79
& 80.26
& 80.33
& 80.79 \\
\hline
\end{tabularx}
\caption{Average STS scores of all-mpnet-base-v2 without and with paraphrase augmentation using Qwen2.5 models with different number of parameters in \%.}
\label{tab:ablation_gen_models}
\end{table}

%
\begin{table}[h]
\small
\centering
\renewcommand{\arraystretch}{1.1} 
\begin{tabular}{lccc}
\hline
\makecell{\textbf{Embedding} \\ \textbf{model}} & \makecell{\textbf{Augmen-} \\ \textbf{tation}} & \textbf{PC} & \textbf{IR} \\
\hline
\multirow{2}{*}{\makecell[l]{glove.840B.300d}}
    & None & 58.41 & 11.83 \\
    & Paraphrase & \textbf{64.64} & \textbf{15.11} (14.12) \\\hline

\multirow{2}{*}{\makecell[l]{bert-large-cased}}
    & None & 55.07 & 4.62 \\
    & Paraphrase & \textbf{62.36} & \textbf{6.66} (5.62) \\\hline

\multirow{2}{*}{\makecell[l]{all-MiniLM-L6-v2}}
    & None & 82.34 & 31.57 \\
    & Paraphrase & \textbf{83.80} & \textbf{33.09} (31.82) \\\hline

\multirow{2}{*}{\makecell[l]{all-mpnet-base-v2}}
    & None & 83.00 & 33.29 \\
    & Paraphrase & \textbf{84.61} & \textbf{35.21} (34.39) \\\hline

\multirow{2}{*}{\makecell[l]{embed-english-\\light-v3.0}}
    & None & 80.18 & 21.56 \\
    & Paraphrase & \textbf{82.88} & \textbf{24.02} (21.63) \\\hline

\multirow{2}{*}{\makecell[l]{embed-english-\\v3.0}}
    & None & 83.54 & \textbf{35.55} \\
    & Paraphrase & \textbf{85.12} & 35.10 (34.34) \\\hline

\multirow{2}{*}{\makecell[l]{voyage-2}}
    & None & 83.81 & 29.65 \\
    & Paraphrase & \textbf{85.24} & \textbf{30.34} (28.96) \\\hline

\multirow{2}{*}{\makecell[l]{voyage-lite-02-\\instruct}}
    & None & \textbf{93.08} & 35.95 \\
    & Paraphrase & 92.22 & \textbf{36.38} (35.50) \\\hline
    
\multirow{2}{*}{\makecell[l]{voyage-large-2}}
    & None & \textbf{92.96} & \textbf{33.20} \\
    & Paraphrase & 92.36 & 32.83 (31.43) \\\hline

\multirow{2}{*}{\makecell[l]{voyage-large-2-\\instruct}}
    & None & 85.21 & \textbf{37.87} \\
    & Paraphrase & \textbf{86.22} & 37.65 (37.44) \\\hline

\multirow{2}{*}{\makecell[l]{mxbai-embed-\\large-v1}}
    & None & 87.16 & 37.73 \\
    & Paraphrase & \textbf{87.38} & \textbf{38.89 }(37.94) \\\hline

\multirow{2}{*}{\makecell[l]{llama-3-8b}}
    & None & 70.10 & 25.39 \\
    & Paraphrase & \textbf{72.78} & \textbf{28.24} (26.39) \\\hline
\end{tabular}
\caption{Avg. Pair Classification (PC) and Information Retrieval (IR) scores with and without paraphrase augmentation (bold: highest scores per embedding model and task). The bracketed IR score shows augmentation applied only to the queries, not the corpus.} 
\label{tab:pc_ir_scores} 
\end{table}
\begin{table}
\small
\renewcommand{\arraystretch}{1.2} 
\centering
\begin{tabular}{lc}\hline
\textbf{Augmentation} & \textbf{all-mpnet-base-v2}  \\\hline
None                                        & 80.28 \\
\makecell[l]{paraphrases + summaries}    & 82.26  \\
\makecell[l]{paraphrases + keywords}     & 82.06  \\
\makecell[l]{summaries + keywords}       & 82.03  \\
\makecell[l]{paraphrases + summa- \\ ries + keywords} & \textbf{82.79}  \\ \hline
\end{tabular}
\caption{Average STS scores for different combinations of augmentations generated with GPT-3.5-Turbo for \texttt{all-mpnet-base-v2} embeddings in \% (bold: highest score).} 
\label{tab:cross_augmentation_scores}
\end{table}
\subsection{Extensions}
To evaluate the generalisability of GASE we extended our work as follows:

\textbf{Multilingual.} Consistent with prior findings, GASE evaluated on the $2$ multilingual MTEB STS datasets containing $12$ different languages using $3$ multilingual embedding models yields greater improvements for weaker encoders, while summarisation remains most effective for BERT and STS22 (\autoref{tab:scores_multilingual}).\footnote{See \autoref{tab:app_sts17_multilingual} and \autoref{tab:app_sts22_multilingual} for results per dataset.}

\mbox{\textbf{Generative model ablation.}} Ablating the generative model size of Qwen 2.5 ~\citep{Qwen25} shows that GASE's performance correlates with generative model size (\mbox{\autoref{tab:ablation_gen_models}}).\footnote{See \mbox{\autoref{sec:appendix-implementation-details}} for implementation details.}

\textbf{Additional task.} We evaluated GASE on Pair Classification (PC) and Information Retrieval (IR). PC was measured with average precision on Twitter SemEval 2015, Twitter URL Corpus, and Sprint Duplicate Questions, while IR used NDCG@10 on NFCorpus. Consistent with STS, GASE improved scores for most models, with gains negatively correlated with baseline performance (\mbox{\autoref{tab:pc_ir_scores}}).\footnote{See \autoref{tab:app_scores_pi_details} in the Appendix for scores per dataset and \citet{MTEB} for details on score calculations.} 

\textbf{Cross-augmentation.} Rather than applying a single augmentation strategy across $k$ different generative models, we used a single generative model with multiple augmentation strategies. Tested with \texttt{all-mpnet-base-v2}, using $3$ augmentations the approach outperformed all other (\autoref{tab:cross_augmentation_scores}).
%
\section{Discussion}%
We believe that embedding models with lower baseline performance benefited more from the semantic diversity induced by augmentation due to their inherent lower capability to model diverse semantics. Similarly, they gained more from the ensemble effect using $k \geq 2$ generative models. Our extension to $11$ non-English languages and the PC and IR tasks did confirm the effectiveness of GASE and also showed a negative association between baseline model performance and improvements through augmentation.
%
%
\begin{figure}[t]
  \centering
  \includegraphics[width=\linewidth]{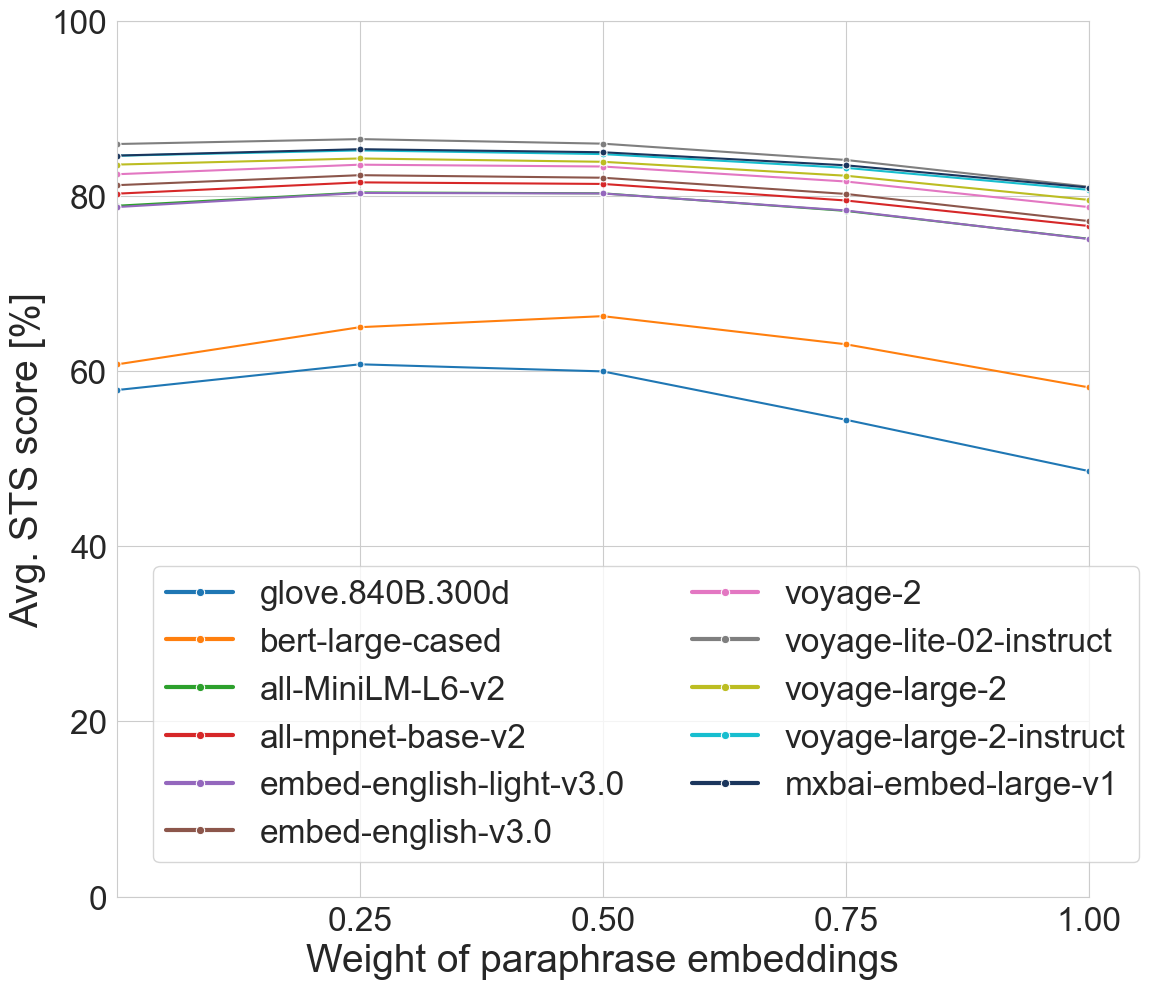}
  \caption{Avg. STS scores for differently weighted averages between embeddings of paraphrases generated with GPT-3.5-Turbo and original texts ($k=1$).}
  \label{fig:weight_analysis}
\end{figure}

The effectiveness of keyword extraction for GloVe, BERT, and Llama-3 may be explained by the model's limited capability to handle complex semantics. Therefore, the reduction to keywords might help the model to reduce noise. 

Longer texts likely caused summary augmentation to outperform paraphrase augmentation on STS22, as summarising significantly reduced the length. On other datasets, however, summaries were only slightly shorter, diminishing its augmentation value. Moreover, the lower Jaccard Similarity between paraphrases and the original texts indicates greater variation, which in turn makes embeddings more effective. Consistent with this, summaries outperform paraphrases on STS22 (\autoref{tab:scores_overview}) - the only dataset where summaries exhibit a lower Jaccard Similarity than paraphrases (\autoref{tab:jaccard_sim}).
Our results show pooling original and synthetic embeddings outperforms individual use (\autoref{fig:weight_analysis}). Across datasets, tasks, and languages, GASE particularly benefits embedding models with lower baseline performance and shorter texts, while summary augmentation is more effective for longer texts (\autoref{tab:scores_overview}). Moreover, GASE requires a certain size of the generative model to be effective (\mbox{\autoref{tab:ablation_gen_models}}). For IR, we found that augmenting both the queries \textit{and} the corpus is necessary (\autoref{tab:pc_ir_scores}) which is particularly costly due to document length.

Since GenEOL used different, mostly weaker, embedding models, our results cannot be directly compared with their results.
\section{Conclusion}
We propose GASE, which augments sentence encoders at inference with generative models for paraphrasing, summarising, or keyword extraction. On MTEB STS, GASE significantly improves lower-performing embedding models and incrementally enhances SOTA models. GASE applies broadly, needing only embedding and generative model outputs. Extensions to $11$ non-English languages and Pair Classification and Information Retrieval as additional tasks, demonstrated consistent findings. 

\section{Limitations}
This study presents several limitations:
\begin{enumerate}
    \item Our approach significantly increases the computational cost for sentence encoding since each generative augmentation requires an additional inference by the embedding model plus inference using a generative LLM. For a basic estimate of the runtime increase based on STS17 see~\autoref{tab:app_runtime}. The results highlight GASE's particular suitability for smaller embedding models, where it yields significant performance gains with minimal encoder runtime.

    \item While deterministic embedding models such as GloVe yield consistent results, other models like GPT-3.5 Turbo examined in this research exhibit stochastic behaviour. Consequently, additional experiments are necessary to corroborate our findings, as LLMs can produce diverse outcomes, thereby limiting the conclusiveness of a single experimental run~\citep{Reimers2018}. This caveat is particularly pertinent to observations based on marginal performance differentials.

    \item Our experiments demonstrate that the effectiveness of GASE is limited to embedding models with lower baseline performance while encoders with higher baseline performance exhibit minimal to no improvement. Further investigation is needed to understand the underlying reasons for this observation.
    
    \item Our investigation was confined to evaluating GPT-3.5 Turbo, Reka-Flash, GPT-4o mini, and to some limited extent Qwen2.5 for generative augmentation. Alternative LLMs, including the Claude and Gemini model families, may lead to different results.

    \item Furthermore, our experimental design was restricted to $k \in \{0,1,2,3\}$. While values of $k>3$ were not explored, we posit that such configurations may be impractical for real-world applications due to prohibitive computational costs.

    \item Our experiments for generative model ablation are limited to models with a parameter count $\in \{0.5b, 1.5b, 3b, 7b, 14b\}$. Extending the analysis to larger generative models would yield insights into whether the observed pattern holds for models with more than $14$b parameters.

    \item Summaries typically comprise $50\%$ or fewer words relative to its source text~\citep{radev2002}. However, due to the concise nature of the texts within the examined STS datasets the summaries generated in this work approximately maintain the original word count. (Except for STS22 which contains longer text sequences.) Hence, the efficacy of augmentation through summarisation may be more pronounced when applied to datasets comprising longer textual inputs (e.g., paragraphs) compared to the predominantly short text sequences examined within the scope of this study. 

\end{enumerate}

\section*{Acknowledgments}
Code development for this work has been assisted by GPT-3.5 Turbo, GPT-4 Omni, Claude 3.5 Sonnet, Gemini Pro 1.5, and Github Copilot.

This research was partially supported by the Horizon Europe project GenDAI (Grant Agreement ID: 101182801) and by the ADAPT Research Centre at Munster Technological University. ADAPT is funded by Taighde Éireann – Research Ireland through the Research Centres Programme and co-funded under the European Regional Development Fund (ERDF) via Grant 13/RC/2106\_P2.

We would also like to thank the anonymous reviewers
for their valuable feedback and constructive suggestions,
which have helped to improve the quality and clarity of this work.

\bibliography{custom}
\appendix
%
%
\section{Comparison of GASE and GenEOL}
A comparison of our approach, GASE, and \mbox{GenEOL} can be found in \mbox{\autoref{tab:app_gase_geneol}}.
\begin{table*}[h]
\small
\centering
\renewcommand{\arraystretch}{1.1}
\begin{tabular}{p{2.5cm}p{4.5cm}p{4.5cm}}
\hline
& \textbf{GASE (ours)} & \textbf{GenEOL} \\ \hline
\textbf{Transformations} 
    & Paraphrasing \newline Summarising \newline Extracting keywords  
    & Paraphrasing \newline Consice paraphrasing \newline Changing the sentence structure \newline Entailment \newline\newline Each one followed by summarisation. \\ \hline
\textbf{Number of transformations} 
    & \mbox{$k\in \{0,1,2,3\}$} 
    & $m\in \{0,2, 4, 8, 16, 24, 32\}$ \\ \hline
\textbf{Pooling} 
    & Mean pooling over variations generated \textit{using $k$ different generative LLMs} 
    & Mean pooling over variations generated \textit{using $m$ different transformations} \\ \hline
\textbf{Embeding models} 
    & English-language models: \newline \texttt{glove.840B.300d} \newline \texttt{bert-large-cased} \newline \texttt{all-MiniLM-L6-v2} \newline \texttt{all-mpnet-base-v2} \texttt{embed-english-light-v3.0} \newline \texttt{embed-english-v3.0} \newline \texttt{voyage-2} \newline \texttt{voyage-lite-02-instruct} \newline \texttt{voyage-large-2} \texttt{voyage-large-2-instruct} \newline \texttt{mxbai-embed-large-v1} \newline \texttt{llama-3-8b} \newline \newline Multilingual models: \newline \texttt{paraphrase-multilingual-\newline mpnet-base-v2} \newline \texttt{multilingual-e5-large-\newline instruct} \newline \texttt{voyage-multilingual-2}
    & \texttt{mistral0.1-7B} \newline \texttt{llama-2-7B} \newline \texttt{llama-3-8B} \newline \texttt{bert-large} \\ \hline
\textbf{Generative models} 
    & \texttt{gpt-3.5-turbo-0125} \newline \texttt{reka-flash-20240226} \newline \texttt{gpt-4o-mini-2024-07-18} \newline \texttt{qwen2.5-0.5b} \newline \texttt{qwen2.5-1.5b}\newline \texttt{qwen2.5-3b} \newline \texttt{qwen2.5-7b} \newline \texttt{qwen2.5-14b} \newline \texttt{qwen2.5-32b} 
    & \texttt{gpt-3.5-turbo-0125} \newline \texttt{mistral0.1-I-7B} \\ \hline
\textbf{STS datasets} 
    & Complete English MTEB for STS: \newline STSB \newline STS12 \newline STS13 \newline STS14 \newline STS15 \newline STS16 \newline STS17 \newline STS22 \newline SICK-R \newline BIOSSES \newline \newline Multilingual datasets: \newline STS17 \newline STS22 
    & Subset of English MTEB for STS: \newline STSB \newline STS12 \newline STS13 \newline STS14 \newline STS15 \newline STS16 \newline SICK-R \\ \hline 
\textbf{Other tasks} 
    & Text Pair Classification \newline Retrieval & Text Pair Classification \newline Classification \newline Clustering \newline  Reranking \\ \hline 
\end{tabular}
\caption{Comparison of GASE and GenEOL.}
\label{tab:app_gase_geneol}
\end{table*}
Based on $7$ STS datasets, \mbox{\citet{Thirukovalluru2024}} report a Spearman's rank correlation for the weakest baseline model of $47.06$ and $80.37$ for their best model using $32$ generative augmentations. Using the same $7$ datasets, we find that \mbox{\textit{without any augmentation}} their best performing model is outperformed by $7$ of the $12$ embedding models evaluated in this work (e.g. by \mbox{\texttt{all-mpnet-base-v2}}; see \mbox{\autoref{tab:app_paraphrasing_per_model_details_1} and \autoref{tab:app_paraphrasing_per_model_details_2}}).
\FloatBarrier
\section{Random keyword extraction}
\label{sec:appendix-random-keyword-extraction}
Random keyword extraction randomly selects $16.1 / 58.2 \approx 0.28$ of the words from a text sequence. This share is equal to the average share of the number of LLM-extracted keywords and the total word count of a sentence (see~\autoref{tab:word_count_augmentation}). As this results in a low number of keywords for short sentences, the minimal number of randomly selected keywords was set to $3$. All punctuation has been removed before extracting the keywords.

\section{Prompts}
\label{sec:appendix-prompts}
Paraphrasing with GPT-3.5 Turbo and GPT-4o mini :\\
\textit{"Rephrase the following text while maintaining its original meaning. If the text contains only a single word, provide a definition or a synomym. When done, check and make sure that the length of the original is approximately maintained. Text:"}
\\

Paraphrasing with Reka-Flash:\\
\textit{"Rephrase the following text while maintaining its original meaning. Do not provide multiple alternatives. Before you reply, remove any explanations. Do only reply with the paraphrased text. If the text contains only a single word, provide a definition or a synomym. Text:"}
\\

Summarising with GPT-3.5 Turbo:\\
\textit{"summarise the following text. Do not include any meta text and only output the summary. If the text is too short to summarise, paraphrase it instead. Text:"}
\\

Extracting keywords with GPT-3.5 Turbo:\\
\textit{"Extract the keywords from the following sentence. Do NOT inlcude any commas or fullstops in your response and do not start your answer with "keywords". Text: "}
\\

Model-specific post-processing is performed to remove meta-text not used for augmentation.

\section{Embeddings models}
\label{sec:appendix-embedding-models}
\textbf{Embedding models for the English language:}
\begin{itemize}
    \item \texttt{glove.840B.300d}~\citep{Pennington2014}
    \item \texttt{bert-large-cased}~\citep{Devlin2019}
    \item \texttt{all-MiniLM-L6-v2}~\cite{SBERT_Original-Models}
    \item \texttt{all-mpnet-base-v2}~\cite{SBERT_Original-Models}
    \item \texttt{embed-english-light-v3.0}~\cite{Cohere_model_overview}
    \item \texttt{embed-english-v3.0}~\cite{Cohere_model_overview}
    \item \texttt{voyage-2}~\citep{VoyageAI_model_overview}
    \item \texttt{voyage-lite-02-instruct}~\citep{VoyageAI_model_overview}
    \item \texttt{voyage-large-2}~\citep{VoyageAI_model_overview}
    \item \texttt{voyage-large-2-instruct}~\citep{VoyageAI_model_overview}
    \item \texttt{mxbai-embed-large-v1}~\cite{Mixedbread_large}
    \item \texttt{llama-3-8b}~\cite{Llama3}
\end{itemize}

For \texttt{llama-3-8b} we used the following KEEOL~\cite{Zhang2024} prompt:
\begin{quote}
    The essence of a sentence is often captured by its main subjects and actions, while descriptive terms provide additional but less central details. With this in mind, this sentence: ``\texttt{input\_text}'' means in one word:
\end{quote}
Following \citet{Zhang2024}, we used the penultimate layer of \texttt{llama-3-8b} to extract the embeddings.
\\

\textbf{Multilingual embedding models:}
\begin{itemize}
    \item \texttt{paraphrase-multilingual-mpnet-\\base-v2}~\cite{Reimers2019}
    \item \texttt{intfloat/multilingual-e5-\\large-instruct}~\cite{intfloat-multilingual-e5-large-instruct}
    \item \texttt{voyage-multilingual-2}~\cite{VoyageAI_model_overview}
\end{itemize}

\section{Datasets}
\label{sec:appendix-datasets}
The following datasets from the MTEB STS leaderboard~\citep{MTEB_leaderboard} were used:

\begin{itemize}
    \item STSB~\citep{SemEval2017_Task1}
    \item STS12~\citep{SemEval2012_Task6}
    \item STS13~\citep{SemEval2013_SharedTask}
    \item STS14~\citep{SemEval2014_Task10}
    \item STS15~\citep{SemEval2015_Task2}
    \item STS16~\citep{SemEval2016_Task1}
    \item STS17~\citep{SemEval2017_Task1}
    \item STS22~\citep{SemEval2022_Task8}
    \item SICK-R~\cite{SICK}
    \item BIOSSES~\citep{BIOSSES}
\end{itemize}

The original MTEB paper~\citep{MTEB} also included the STS11 dataset in Figure 1 of their paper where the authors specify the MTEB datasets, but excluded it in the evaluations. Similarly, the MTEB Leaderboard~\citep{MTEB_leaderboard} does not include STS11 either. To make our results comparable to other models evaluated on MTEB we therefore excluded STS11 from our evaluations. All data has been obtained from the MTEB repository on the Hugging Face website.\footnote{\url{https://huggingface.co/mteb/datasets}}
\\

The following datasets were used for our extensions:
\begin{itemize}
    \item Twitter SemEval 2015~\citep{twitter-semeval-2015}
    \item Twitter URL-Corpus~\citep{TwitterURLCorpus}
    \item Sprint duplicate questions~\citep{SprintDuplicateQuestions}
    \item NFCorpus~\citep{Boteva2016}
\end{itemize}

\section{Generative Model Versions, Hyperparameters, and Hardware Configuration}
\label{sec:appendix-implementation-details}
The following versions have been used for generative augmentation:
\begin{itemize}
    \item GPT-3.5 Turbo: \texttt{gpt-3.5-turbo-0125}
    \item Reka-Flash: \texttt{reka-flash-20240226}
    \item GPT-4o mini: \texttt{gpt-4o-mini-2024-07-18}
    \item Qwen2.5 (all models running via Ollama with quantization Q4\_K\_M):
        \begin{itemize}
            \item \texttt{qwen2.5:0.5b} 
            \item \texttt{qwen2.5:1.5b}
            \item \texttt{qwen2.5:3b}
            \item \texttt{qwen2.5:7b}
            \item \texttt{qwen2.5:14b}
        \end{itemize}

\end{itemize}

GPT-3.5 Turbo, Reka-Flash, and GPT-4o mini have been accessed through the respective APIs with \mbox{\texttt{temperature}} set to $0$ and \mbox{\texttt{top\_p}} to $1$ to make the experiments as reproducible as possible. Where applicable, a random seed of $1337$ was used. Other hyperparameters were used with default values. 

All Qwen2.5 model runs were conducted with Ollama on a local computer with \texttt{Q4\_K\_M} quantisation and the following specification:
\begin{itemize}
    \item GeForce RTX 3090 MSI Gaming X Trio 24GB GPU
    \item AMD Ryzen 9 9950X CPU
    \item DDR5-6400 64GB RAM
\end{itemize}

\section{Additional Experimental Results}
\label{sec:appendix-results}
\begin{table}[ht]
\small
\centering
\renewcommand{\arraystretch}{1.1} 
\begin{tabular}{lcc} 
\hline 
\textbf{Embedding model} & \textbf{Pooling} & \textbf{STS Score} \\ 
\hline 
\multirow{2}{*}{\makecell[l]{glove.840B.300d}}
    & mean & \textbf{59.98} \\ 
    & max & 59.02 \\\hline 
\multirow{2}{*}{\makecell[l]{bert-large-cased}}
    & mean & \textbf{66.29 }\\ 
    & max & 65.16 \\\hline 

\multirow{2}{*}{\makecell[l]{all-MiniLM-L6-v2}}
    & mean & \textbf{80.32} \\ 
    & max & 79.88 \\\hline 

\multirow{2}{*}{\makecell[l]{all-mpnet-base-v2}}
    & mean & \textbf{81.39} \\ 
    & max & 81.03 \\\hline 

\multirow{2}{*}{\makecell[l]{embed-english-\\light-v3.0}}
    & mean & \textbf{80.30} \\ 
    & max & 79.74 \\\hline 

\multirow{2}{*}{\makecell[l]{embed-english-v3.0}}
    & mean & \textbf{82.10} \\ 
    & max & 81.62 \\\hline 

\multirow{2}{*}{\makecell[l]{voyage-2}}
    & mean & \textbf{83.38} \\ 
    & max & 82.69 \\\hline 

\multirow{2}{*}{\makecell[l]{voyage-lite-02-\\instruct}} 
    & mean & \textbf{85.99} \\ 
    & max & 85.53 \\\hline 
    
\multirow{2}{*}{\makecell[l]{voyage-large-2}}
    & mean & \textbf{83.92} \\ 
    & max & 83.48 \\\hline 

\multirow{2}{*}{\makecell[l]{voyage-large-2-\\instruct}} 
    & mean & \textbf{84.80} \\ 
    & max & 84.41 \\\hline 

\multirow{2}{*}{\makecell[l]{mxbai-embed-\\large-v1}} 
    & mean & \textbf{85.01} \\ 
    & max & 84.69 \\\hline 

\multirow{2}{*}{\makecell[l]{llama-3-8b}} 
    & mean & \textbf{75.28} \\ 
    & max & 75.07 \\\hline 
\end{tabular}
\caption{Average STS scores across tasks for different pooling methods in \% (bold: best per embedding model).}
\label{tab:pooling_scores_avg} 
\end{table}

%
\begin{table*}[ht]
\tiny
\renewcommand{\arraystretch}{1.2} 
\centering
\begin{tabular}{l l l c c c c c c c c c c}\hline
\textbf{Embedding  model}              & \textbf{Augmentation} & \textbf{Average} & \textbf{STSB}  & \textbf{STS12} & \textbf{STS13} & \textbf{STS14} & \textbf{STS15} & \textbf{STS16} & \textbf{STS17} & \textbf{STS22} & \textbf{SICK-R} & \textbf{BIOSSES} \\\hline
\multirow{4}{*}{glove.840B.300d}           & none                      & 57.86          & 50.73          & 57.50          & 70.98          & 60.69          & 70.85          & 63.85          & 62.05          & 54.08          & 55.42           & 32.45            \\
                                           & paraphrasing           & 59.98          & 56.78          & 57.33          & \textbf{71.17} & 61.16          & 71.52          & \textbf{67.66} & 62.28          & 57.78          & 58.92           & \textbf{35.21}   \\
                                           & summarising           & 60.64          & 57.67          & 55.00          & 70.24          & \textbf{61.18} & 71.64          & 66.81          & 67.08          & \textbf{61.21} & \textbf{60.92}  & 34.65            \\
                                           & extracting keywords   & \textbf{60.82} & \textbf{59.26} & \textbf{58.09} & 71.12          & 60.34          & \textbf{72.44} & 65.31          & \textbf{75.92} & 56.94          & 55.34           & 33.50            \\\hline
\multirow{4}{*}{bert-large-cased}          & none                      & 60.78          & 59.18          & 45.71          & 63.94          & 54.42          & 68.98          & 65.46          & 71.30          & 55.48          & 64.16           & 59.14            \\
                                           & paraphrasing           & 66.29          & 67.97          & \textbf{53.62} & 68.56          & 59.73          & 74.01          & \textbf{71.97} & 74.14          & 58.57          & 69.79           & \textbf{64.56}   \\
\hline

\end{tabular}
\caption{Scores per dataset with and without generative augmentation using GPT-3.5 Turbo in \% (bold: highest scores).}
\label{tab:app_scores_per_augm_details}
\end{table*}

%
\begin{table*}
\tiny
\renewcommand{\arraystretch}{1.2} 
\centering
%
\caption{Scores per dataset with and without stopword removal using either no generative augmentation or paraphrase augmentation with GPT-3.5 Turbo in \%.} 
\label{tab:app_scores_stopword}
\end{table*}

\begin{table*}
\tiny
\centering
\renewcommand{\arraystretch}{1.2} 
\begin{tabular}{@{}l l c c c c c c c c c c c@{}} 
\hline
\textbf{Embedding model}              & \textbf{Gen model}                            & \textbf{Average} & \textbf{STSB}  & \textbf{STS12} & \textbf{STS13} & \textbf{STS14} & \textbf{STS15} & \textbf{STS16} & \textbf{STS17} & \textbf{STS22} & \textbf{SICK-R} & \textbf{BIOSSES} \\ \hline
\multirow{6}{*}{glove.840B.300d}      & none                                             & 57.86            & 50.73          & 57.50          & 70.98          & 60.69          & 70.85          & 63.85          & 62.05          & 54.08          & 55.42           & 32.45            \\
                                      & gpt-3.5-turbo                                 & 59.98            & 56.78          & 57.33          & 71.17          & 61.16          & 71.52          & \textbf{67.66} & 62.28          & 57.78          & 58.92           & 35.21            \\
                                      & reka-flash                                    & 60.73            & 55.02          & 59.53          & 71.09          & 61.22          & 72.81          & 65.20          & 64.27          & 60.11          & 61.65           & 36.43            \\
                                      & gpt-4o-mini                                   & 60.04            & 55.09          & 60.12          & 72.65          & 62.45          & 71.60          & 66.49          & 63.39          & 55.48          & 60.36           & 32.76            \\
                                      & \makecell[l]{gpt-3.5-turbo\\+ reka-flash}     & 61.97            & 59.46          & 60.06          & 72.01          & 62.68          & 72.46          & 67.62          & \textbf{65.43} & 61.46          & 61.51           & \textbf{36.95}   \\
\hline
\end{tabular}
\caption{Scores per dataset with and without paraphrase augmentation using GPT-3.5 Turbo, Reka-Flash, and GPT-4o mini in \% (bold: highest scores) (part 1).} 
\label{tab:app_paraphrasing_per_model_details_1} 
\end{table*}
%
\begin{table*}
\tiny
\centering
\renewcommand{\arraystretch}{1.2} 
\begin{tabular}{@{}l l c c c c c c c c c c c@{}} 
\hline
\textbf{Embedding model}              & \textbf{Gen model}                            & \textbf{Average} & \textbf{STSB}  & \textbf{STS12} & \textbf{STS13} & \textbf{STS14} & \textbf{STS15} & \textbf{STS16} & \textbf{STS17} & \textbf{STS22} & \textbf{SICK-R} & \textbf{BIOSSES} \\ \hline
\multirow{6}{*}{voyage-2}             & none                                             & 82.50            & 87.06          & 77.68          & 86.25          & 80.11          & 88.38          & 85.72          & \textbf{89.78} & 65.26          & 78.95           & 85.83            \\
                                      & gpt-3.5-turbo                                 & 83.38            & 87.17          & 79.23          & 87.13          & 81.67          & 88.57          & 86.27          & 89.07          & 66.80          & 80.33           & \textbf{87.56}   \\
                                      & reka-flash                                    & 83.37            & 86.93          & 79.65          & 87.83          & 81.79          & 88.48          & 86.18          & 88.98          & 67.42          & 80.67           & 85.72            \\
                                      & gpt-4o-mini                                   & 83.07            & 86.89          & 79.59          & 87.27          & 81.23          & 88.46          & 85.81          & 89.37          & 65.25          & 80.85           & 86.02            \\
                                      & \makecell[l]{gpt-3.5-turbo\\+ reka-flash}     & \textbf{83.80}   & \textbf{87.27} & 80.62          & 87.99          & 82.40          & 88.51          & \textbf{86.40} & 88.77          & \textbf{67.94} & 80.87           & 87.28            \\
\hline
\end{tabular}
\caption{Scores per dataset with and without paraphrase augmentation using GPT-3.5 Turbo, Reka-Flash, and GPT-4o mini in \% (bold: highest scores) (part 2).}
\label{tab:app_paraphrasing_per_model_details_2} 
\end{table*}

\begin{table*}[htbp] 
\small 
\centering
\renewcommand{\arraystretch}{1.2} 
\begin{tabular}{lcccccc}
\hline
\multirow{2}{*}{\textbf{Dataset}} & \multicolumn{2}{c}{\textbf{\makecell[c]{paraphrase-multilingual-\\mpnet-base-v2}}} & \multicolumn{2}{c}{\textbf{\makecell[c]{multilingual-\\e5-large-instruct}}} & \multicolumn{2}{c}{\textbf{\makecell[c]{voyage-multilingual-2}}} \\
\cline{2-7}
 & None & \makecell{Paraphrase\\ Augmentation} & None & \makecell{Paraphrase\\ Augmentation} & None & \makecell{Paraphrase\\ Augmentation} \\
\hline
STS17\_en-en & \textbf{86.99} & 86.01          & \textbf{87.64} & 87.32          & \textbf{90.81} & 90.20          \\
STS17\_ar-ar & 79.09          & \textbf{81.04} & 81.61          & \textbf{83.36} & 80.53          & \textbf{81.58} \\
STS17\_en-ar & 80.85          & \textbf{81.76} & 78.66          & \textbf{79.50} & 81.59          & \textbf{83.48} \\
STS17\_en-de & 83.28          & \textbf{83.81} & \textbf{85.09} & 85.05          & 88.09          & \textbf{88.11} \\
STS17\_en-tr & 74.90          & \textbf{75.44} & 77.04          & \textbf{78.88} & 74.12          & \textbf{77.09} \\
STS17\_es-en & 86.11          & \textbf{86.25} & 84.72          & \textbf{85.35} & \textbf{88.24} & 87.74          \\
STS17\_es-es & \textbf{85.14} & 85.00          & \textbf{88.09} & 87.87          & \textbf{88.71} & 88.63          \\
STS17\_fr-en & 81.17          & \textbf{82.47} & 83.26          & \textbf{84.09} & 87.10          & \textbf{87.65} \\
STS17\_it-en & \textbf{84.24} & 84.20          & 84.94          & \textbf{85.00} & 88.98          & \textbf{89.43} \\
STS17\_ko-ko & 83.29          & \textbf{83.83} & 81.58          & \textbf{82.47} & 80.17          & \textbf{80.69} \\
STS17\_nl-en & 82.51          & \textbf{82.65} & \textbf{84.65} & 84.59          & 88.22          & \textbf{88.38} \\
\hline
\end{tabular}
\caption{Multilingual STS17 Scores with and without paraphrase augmentation using GPT-4o mini in \% (bold: highest score per embedding model and dataset). }
\label{tab:app_sts17_multilingual}
\end{table*}

\begin{table*}[htbp] 
\small 
\centering
\renewcommand{\arraystretch}{1.2} 
\begin{tabular}{lcccccc}
\hline
\multirow{2}{*}{\textbf{Dataset}} & \multicolumn{2}{c}{\textbf{\makecell[c]{paraphrase-multilingual-\\mpnet-base-v2}}} & \multicolumn{2}{c}{\textbf{\makecell[c]{multilingual-e5-\\large-instruct}}} & \multicolumn{2}{c}{\textbf{\makecell[c]{voyage-multilingual-2}}} \\
\cline{2-7}
 & None & \makecell{Paraphrase\\ Augmentation} & None & \makecell{Paraphrase\\ Augmentation} & None & \makecell{Paraphrase\\ Augmentation} \\
\hline
STS\_ar-ar   & \textbf{64.77} & 56.36          & \textbf{81.61} & 65.02          & 59.65          & \textbf{60.19} \\
STS\_de-de   & \textbf{62.30} & 48.16          & \textbf{78.66} & 62.00          & 63.01          & \textbf{63.08} \\
STS\_de-en   & \textbf{62.85} & 57.03          & \textbf{85.09} & 61.72          & \textbf{62.87} & 62.56          \\
STS\_de-fr   & \textbf{64.19} & 62.68          & \textbf{77.04} & 65.52          & \textbf{64.88} & 64.12          \\
STS\_de-pl   & \textbf{56.01} & 41.26          & \textbf{84.72} & 56.97          & 54.14          & \textbf{56.35} \\
STS\_es-es   & \textbf{73.27} & 65.03          & \textbf{88.09} & 73.07          & 75.32          & \textbf{75.87} \\
\hline
\end{tabular}
\caption{Multilingual STS22 Scores with and without paraphrase augmentation using GPT-4o mini in \% (bold: highest score per embedding model and dataset). }
\label{tab:app_sts22_multilingual}
\end{table*}

%
\begin{table*}[ht]
\small
\centering
\renewcommand{\arraystretch}{1.1} 
\begin{tabular}{lcccc}
\hline 
\textbf{Embedding model} & \textbf{\makecell{Augmen-\\tation}} & \multicolumn{3}{c}{\textbf{Text Pair Classification}} \\
 & & \makecell{Twitter Sem-\\Eval 2015} & \makecell{Twitter URL-\\Corpus} & \makecell{Sprint dupli-\\cate questions} \\
\hline 

\multirow{2}{*}{\makecell[l]{glove.840B.300d}}
    & none & 43.81          & 63.09          & 68.33          \\
    & paraphrase & \textbf{50.04} & \textbf{70.49} & \textbf{73.39} \\
\hline 

\multirow{2}{*}{\makecell[l]{bert-large-cased}}
    & none & 47.27          & 69.81          & 48.15          \\
    & paraphrase & \textbf{57.21} & \textbf{76.28} & \textbf{53.61} \\
\hline

\multirow{2}{*}{\makecell[l]{all-MiniLM-L6-v2}}
    & none & 67.86          & 84.61          & 94.55          \\
    & paraphrase & \textbf{71.50} & \textbf{84.70} & \textbf{95.20} \\
\hline 
\end{tabular}
\caption{Performance on Text Pair Classification with and without paraphrase augmentation using GPT-3.5-Turbo in \%.}
\label{tab:app_scores_pi_details}
\end{table*}
\FloatBarrier
\section{Runtime analysis}
\label{sec:app_runtime}
\begin{table*}[hb]
\small
\centering
\renewcommand{\arraystretch}{1.1}
\begin{tabular}{lll}
\hline
\textbf{Augmentation} & \textbf{Average runtime {[}sec{]}} & \textbf{Stand. deviation of runtime {[}sec{]}} \\
\hline
None                  & 2.30                               & 0.40                                           \\
paraphrase            & 132.57                             & 15.87                                          \\
summarise             & 140.48                             & 14.15                                          \\
keywords              & 127.17                             & 18.16                                          \\
\hline
\end{tabular}
\caption{Average runtime and corresponding standard deviation without and with generative augmentation using GPT-3.5-Turbo and, as an encoder, all-mpnet-base-v2 on STS17 (based on $5$ runs each, in seconds).}
\label{tab:app_runtime}
\end{table*}
To provide a basic estimate of the additional runtime introduced by GASE over a non-augmented baseline, we performed five runs using GPT-3.5-Turbo for generation and all-mpnet-base-v2 for embeddings and report the mean and standard deviation. Because the difference in runtimes between the augmented and non-augmented setups was substantial and the standard deviation across runs was small, we did not carry out statistical significance testing (see \autoref{tab:app_runtime}).

All runtime experiments were conducted using Ollama on a local computer with the following specification: GeForce RTX 3090 MSI Gaming X Trio 24GB GPU, AMD Ryzen 9 9950X CPU, DDR5-6400 64GB RAM.
\FloatBarrier
\end{document}